\documentclass{egpubl}
\usepackage{egweblnk}

\JournalSubmission    
\usepackage[T1]{fontenc}

\biberVersion
\BibtexOrBiblatex
\usepackage[backend=biber,bibstyle=EG,citestyle=alphabetic,backref=true]{biblatex} 
\electronicVersion
\PrintedOrElectronic
\ifpdf \usepackage[pdftex]{graphicx} \pdfcompresslevel=9
\else \usepackage[dvips]{graphicx} \fi

\usepackage{amsmath,amssymb,cases} 
\usepackage[labelfont=bf,textfont=it]{caption}
\usepackage[labelfont=bf,textfont=it]{subcaption}
\usepackage{multirow}
\usepackage{graphicx}
\usepackage{mathtools}
\usepackage{xcolor}
\usepackage{tabularx}
\usepackage{booktabs}
\newcolumntype{Y}{>{\centering\arraybackslash}X}

\newcommand\blfootnote[1]{%
  \begingroup
  \renewcommand\thefootnote{\fnsymbol{footnote}}%
  \footnotetext[1]{#1}%
  \endgroup
}

\title{Deep Learning-Based Facial Retargeting Using Local Patches}

\author[Choi et al.]
{\parbox{\textwidth}{\centering 
    Yeonsoo Choi$^{2,*}$,
    Inyup Lee$^{1,*}$,
    Sihun Cha$^{1}$,
    Seonghyeon Kim$^{3}$,
    Sunjin Jung$^{1}$,
    Junyong Noh$^{1}$\orcid{0000-0003-1925-3326}
    }
    \\
    {\parbox{\textwidth}{
    \centering $^1$Visual Media Lab, KAIST, Republic of Korea \\
    \centering $^2$Netmarble F\&C, Republic of Korea \\
    \centering $^3$Anigma, Republic of Korea \\
    }}
}

\begin{document}

\maketitle

\begin{abstract}
In the era of digital animation, the quest to produce lifelike facial animations for virtual characters has led to the development of various retargeting methods. 
While the retargeting facial motion between models of similar shapes have been very successful, challenges arise when the retargeting is performed on stylized or exaggerated 3D characters that deviate significantly from human facial structures.
In this scenario, it is important to consider the target character's facial structure and possible range of motion to preserve the semantics assumed by the original facial motions after the retargeting. To achieve this, we propose a local patch-based retargeting method that transfers facial animations captured in a source performance video to a target stylized 3D character. 
Our method consists of three modules. The Automatic Patch Extraction Module extracts local patches from the source video frame.
These patches are processed through the Reenactment Module to generate correspondingly reenacted target local patches. The Weight Estimation Module calculates the animation parameters for the target character at every frame for the creation of a complete facial animation sequence.
Extensive experiments demonstrate that our method can successfully transfer the semantic meaning of source facial expressions to stylized characters with considerable variations in facial feature proportion.


\end{abstract} 


\section{Introduction}
\blfootnote{*These authors contributed equally to this work.}
Since the release of the first 3D feature animation~\cite{toystory}, acquiring perceptually plausible facial animation for a 3D character has been an important topic in computer graphics. Facial expressions involve delicate and complex movements of underlying muscles. These subtle movements are particularly challenging to replicate on 3D models. Therefore, a wide range of techniques have been attempted in an effort to overcome this difficulty and produce facial animation with great realism.

Keyframing has been a popular choice to produce realistic facial animation, but it requires laborious manual work even for skilled animators.
Therefore, methods based on motion capture were proposed as a viable alternative to obtain realistic facial motion and reproduce it on a target model.
Because of the typical shape difference between the performer and the target model, the utilization of motion capture led to the development of facial animation retargeting~\cite{choe2001performance,deng2006animating,bickel2007multi,seol2012spacetime}. 
Some methods automated this retargeting process by assuming that the appearance of the human performer and that of the target face model is similar to each other~\cite{weise2009face,weise2011realtime,cao20133d,cao2014displaced}.
Because this assumption does not apply to stylized 3D characters, retargeting to a stylized 3D character often accompanies an explicit specification of the deformation relationship between the corresponding points of the two models in the preprocessing step~\cite{noh2001expression,sumner2004deformation,song2011characteristic,seol2012spacetime,saito2013smooth,seol2016creating,onizuka2019landmark}.

Advances in deep learning technologies have made significant breakthroughs in facial animation retargeting, enabling to estimate blendshape or rig parameters for the 3D character using deep neural networks~\cite{costigan2014facial, aneja2018learning, aneja2019learning, chaudhuri2019joint}.
Recent studies utilize an image-based reenactment process, which simply requires an RGB video of a human performer~\cite{kim2021deep, moser2021semi}, eliminating the need for markers or any traditional correspondence configuration between the source and target. This approach is not confined to any specific rig setups, whether blendshape-based or joint-based. 
These deep learning-based approaches also assume that the source and target facial shapes are reasonably similar for the image-based reenactment to be successfully applied.
Any differences in size or location of the face element between the source and the target would cause misalignment in the image space, producing an error in the reenactment process. This error will propagate to the animation parameter computation step, leading to an implausible retargeting result as shown in Figure~\ref{fig:comparison} and Figure~\ref{fig:comparison2}.
To address this, we propose a local patch-based retargeting method that enables to retarget the facial animation of a source human performer to a target stylized 3D character, by aligning each face element in the image space.

The facial action coding system (FACS) \cite{Ekman_1978_10190} decomposes facial expressions into numerous Action Units (AUs). Because AUs are independent local patterns that are combined to generate facial expressions~\cite{deng2008computer}, we decompose the face of the human performer and that of the stylized 3D character into local patches. The retargeting of the facial animation from the source to the target is then performed through the semantically corresponding local patches between them. This patch-based retargeting allows capturing the semantic meaning of the facial expression locally from the human performer. By focusing on local expressions observed around each face element, our method faithfully reproduces the facial expression of the human performer in a source video on the target stylized 3D character.

Our method consists of three modules; \textit{Automatic Patch Extraction Module} (APEM), \textit{Reenactment Module} (RM), and \textit{Weight Estimation Module} (WEM). APEM is a novel local patch extraction technique that ensures precise alignment between source and target faces using a patch-based approach, even with significant shape differences in face elements. During training, APEM extracts local patches from both source video frames and rendered target animation frames. At inference, it solely extracts local patches from source video frames.
After extracting the patches, RM transfers the facial expressions from each source patch to the corresponding target patch, generating matching reenacted target local patches.
WEM then regresses the animation parameters for the target character given the reenacted target local patches. Applying this process at each frame of the source video produces the full facial animation for the target character.

The contributions of the proposed method can be summarized as follows:
\begin{itemize}
    \item Retargeting facial animation to stylized 3D characters whose face elements have significantly different shapes and sizes from those of a human performer.  
    \item A method for transferring facial expressions from a source performance video to a target stylized 3D character model using automatically extracted local patches.
    \item Effectively preserving the semantics of the source facial expressions after the transfer, fully utilizing the target character's range of motion.
\end{itemize}


\section{Related Work}

\subsection{Facial Motion Retargeting}\label{relatedwork_facialretarget} 
Facial motion retargeting is the process of transferring the facial performance captured from a source model to a target model while preserving the semantic meaning of the facial expressions~\cite{pighin2006facial}. 
The goal is to create a visually appealing and perceptually plausible animation that accurately reflects the intended expression of the source performance to the target model.
One way to achieve this is through \textit{parallel parameterization}.
This technique allows the two faces to share a semantically identical expression parameter space~\cite{ribera2017facial}.
Noh and Neumann~\cite{noh2001expression} proposed a method called expression cloning which utilizes Radial Basis Function (RBF) to project the motion vectors from a source face to a target face. 
Sumner and Popovi{\'c}~\cite{sumner2004deformation} suggested a deformation transfer method that copies the deformations exhibited by a source mesh to a target with different mesh structures. Xu et al.~\cite{Xu2014controllable} utilized deformation transfer as well, and also enabled both large and fine-scale editing assuming an identical mesh structure between source and target.
This work was later improved by Gao et al.~\cite{gao2018automatic}, which solves the deformation transfer problem in an unsupervised way using a mesh-based autoencoder structure. 

Other approaches have further explored RBF-based regression techniques using manual expression pairs~\cite{deng2006animating, song2011characteristic} or manually constructed corresponding points~\cite{seol2012spacetime, ribera2017facial}.
Recent advancements in augmented reality frameworks have also influenced facial retargeting methodologies. Apple's augmented reality framework, ARKit~\cite{arkit}, enabled iOS devices to capture facial expressions without using explicit markers, tracking facial movements in real-time using only the camera and sensors embedded in the device.
Different from these previous facial retargeting methods, our approach reduces the manual work, by requiring the selection of a few key points on the target model corresponding to auto-detected facial landmarks on the performer in the source video. This is notably simpler compared to the task of selecting comprehensive corresponding points on both source and target across entire face. 
Specifically, unlike ARKit, which requires manual data preparation or blendshape correspondence, our approach offers a rig-agnostic and mesh-agnostic retargeting method for stylized character animation, automatically learning the correspondence between source and target representations.

\subsection{Video-based Face Retargeting}

With the advance of deep learning, face reenactment and face swapping, commonly referred to as \textit{deepfakes} have gained much attention due to their ability to generate photo-realistic videos with manipulated facial expressions. In face reenactment, the performance from a source video is transferred to another video~\cite{nirkin2019fsgan, thies2020face2face, kim2018deep}. Face swapping, on the other hand, replaces the face in a target video with the face from a source video ~\cite{perov2021deepfacelab, naruniec2020high}.
While they are performed in a 2D video domain, facial retargeting is closely connected to these tasks because it involves mapping facial expressions from source to target.
Several methods have been explored for facial retargeting~\cite{kim2021deep, moser2021semi} to a 3D geometry domain. These methods reenact the face in the 2D image followed by regressing the 3D animation parameters for character-to-character retargeting~\cite{kim2021deep} or human performer to performer-specific/alternate character retargeting~\cite{moser2021semi}. By leveraging image-based approaches, these methods enable the retargeting process without relying on paired datasets, or the need for geometric priors. 
Recently, a keypoint-based approach such as LivePortrait~\cite{guo2024liveportrait} has been proposed to generate high-quality 2D human portrait videos. Unlike the 2D based approach that focuses on real human faces, our method provides explicit animation parameters for 3D characters, ensuring the faithful creation of nuanced expressions while simplifying the animation process without the need for handling of rig controls or predefined correspondences.
Our work extends these 2D image-based techniques by enabling retargeting to stylized 3D characters, focusing on individual face elements through the use of local patches.

\subsection{Facial Retargeting for Stylized Character}

Several studies aim to retarget facial animation to stylized characters. 
Aneja et al.~\cite{aneja2016modeling} trained independent convolutional neural networks for human and stylized characters and used a perceptual model to retrieve facial expression images of a 2D character.
The approach has been extended to generate the expression of a stylized 3D character driven by the image of a human using a neural network that optimizes over expression clarity \cite{aneja2019learning}. 
Different from these methods that require a manually labeled dataset of images and expression labels, our approach does not require a paired dataset.
Zhang et al.~\cite{zhang2020facial} trained variational autoencoders for the human and stylized character to establish a shared latent space between them. Our method also takes a similar approach of establishing a latent space between each face element of the source and target models ~\cite{bouaziz2014semi}.

\begin{figure*}[ht]
 \includegraphics[width=\linewidth]{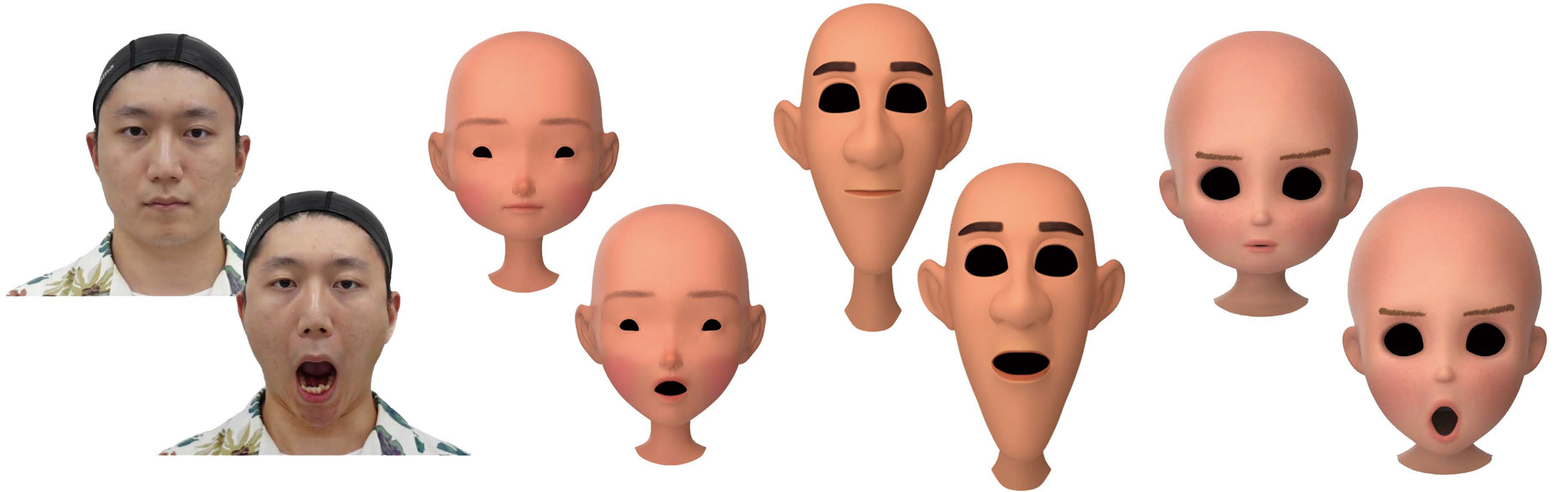}
 \centering
 \caption{
    Our method retargets facial animations from a source performance video to a stylized target 3D character using local patches.
 }
\label{fig:teaser}
\end{figure*}

\begin{figure*}[ht]
    \centerline{\includegraphics[width=0.95\linewidth]{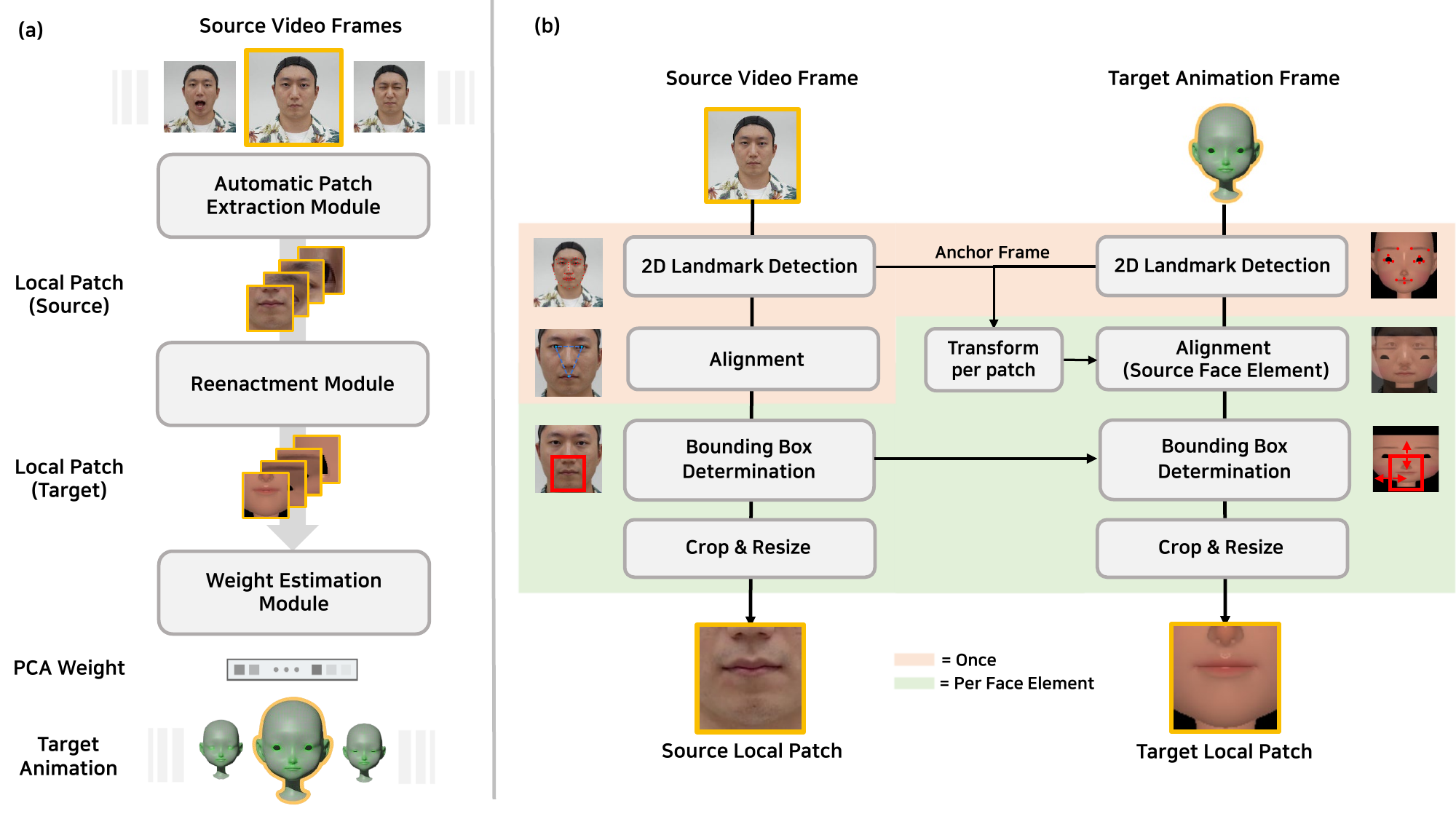}}
    \caption{
    Overview of the proposed method at inference time (a) and the illustration of automatic patch extraction (b) using local patch extraction around the lips as an example.
    } 
    \label{fig:infer3}
\end{figure*}

\subsection{Local Patch-based Approaches}

The approaches that utilize local patches have been explored to address various facial manipulation tasks. 
Tena et al.~\cite{tena2011interactive} introduced a local model for face reconstruction, which achieves more flexibility over the holistic approach due to the use of localized deformation space. Wu et al.~\cite{wu2016anatomically} further improved the robustness of the method by enforcing the patch motion to follow anatomical constraints. 
To produce a high-fidelity model, a dual-pathway network that balances between the global and local models was proposed by Tran et al.~\cite{tran2019towards}. This method enables the reconstruction of a detailed 3D face model from an image. Similarly, Huang et al. \cite {huang2017beyond} introduced a two-pathway network in which the global structure and local details are perceived simultaneously to synthesize a photorealistic frontal view. Chandran et al.~\cite{chandran2020attention} utilize a local patch-based model in landmark detection tasks. They extract the semantic regions of the face to train regional networks using high-resolution images.
Similar to these local patch-based methods, we extract the local patches from the face and align them to construct a model that consists of independently trained networks focusing on specific local patches. By paying attention to individual face elements, our model improves the performance of the task specifically related to those face elements.

Some approaches employ the concept of local retargeting to address a similar objective to our work.
Liu et al.~\cite{liu2011framework} use automatically generated segmentation masks that divide the face into regions to optimize the best blendshape weights for each local patch. The number of generated regions vary depending on the threshold value that determines the number of vertices for each region. 
Chandran et al.~\cite{chandran2022local} perform patch-wise retargeting before applying the result to an anatomical model. This patch-based method enables retargeting between two geometric models of different mesh structures as long as the layouts of the patch are the same between the source and target.
We extract the patches directly from images, avoiding the need for a geometric model for the source and a complex setting.

\section{Method}
In this section, we present the intricacies of the proposed method along with the necessary data inputs.
Figure~\ref{fig:infer3} (a) provides an overview of our method during the inference stage.
The process begins with APEM, which autonomously extracts local patches of the source. Once these local patches are extracted, RM steps in to generate the equivalently reenacted patches for the target character. Following reenactment, these target patches are processed by WEM, which produces Principal Component Analysis space(PCA) weights that will be applied to PCA blendshapes to produce the facial expression of the target 3D character. The process is executed frame by frame, resulting in a complete facial animation. The details of APEM, and training RM and WEM are described in Section~\ref{AutoPatch}, Section~\ref{REM}, and Section~\ref{WEM}, respectively.

\subsection{Data Preparation} \label{data preparation}

To train the networks within our pipeline, pre-processed video frames of a human performer and rendered frames from the 3D animation of a target stylized character, along with their corresponding animation parameters are required. A key aspect of our method lies in its ability to be trained without direct supervision, allowing the source video frames and the target character's animation frames to be unpaired. In the training phase, we utilize image frames and their corresponding 2D landmarks from both the source video and the rendered 3D animation of the target character. Additionally, PCA weights from the target character’s animation are utilized. In the inference phase, only the image frames from the source video are required. The subsequent sections detail the specifics of this data preparation process.

\subsubsection{Video Data}\label{video_data}

For the source, we captured videos of two human performers (one female, one male) exhibiting a broad spectrum of facial movements. Starting from a neutral expression, expressions activated based on FACS were sequentially recorded using a budget-friendly camera at $60$ FPS, totaling $33,618$ frames: $12,896$ from the first performer and $20,722$ from the second. For testing purposes, we reserved $2,122$ and $3,482$ frames from the first and second performers, respectively, with the remaining frames allocated for training. An additional $1,353$ frames were captured from the second performer under varying lighting conditions to assess light condition robustness in Section~\ref{eval_color}. We employed the Phong illumination model~\cite{phong1975illumination} for rendering of the target animation, using orthogonal projection with a camera orientation that closely matches the human performance footage.

\subsubsection{3D Animation Data} 

Our 3D animation data originate from two primary sources: retargeted animations via ARkit and manually crafted animations by an artist. These animations undergo decomposition through PCA. The resulting dimensionality of the PCA parameter vectors is 20 for all experiments, with the basis vectors capturing $99.9\%$ of the total variance. The PCA basis thus functions as a linear blendshape system of the captured data. For each animation frame, we store the PCA weights alongside the rendered local patches. This data is used for training WEM. Further details about the specifics of the 3D animation data used in training are discussed in Section~\ref{implement}.

\subsubsection{Image Augmentation}
During the training, color augmentation and geometry augmentation are applied to the local patches. 
For the color augmentation, we perform color jitter in the range of ($0.5$, $2$) for brightness, ($0.5$, $2$) for saturation, ($0.5$, $2$) for contrast, and ($-0.2$, $0.2$) for hue. For the geometry augmentation, we apply random affine transformations as suggested in Moser et al.~\cite{moser2021semi}: augmentation in the range of ($0^{\circ}$, $10^{\circ}$) for rotation, ($0\%$, $5\%$) for translation, and ($0\%$, $5\%$) for scale.

\subsection{Automatic Patch Extraction Module}\label{AutoPatch}

APEM plays an important role in aligning face elements between source and target characters, a key aspect in the transfer of motion between faces with significant differences in shape and size. The module comprises four stages: 2D landmark detection, alignment, bounding box determination, and cropping, each tailored to the source video frames and the rendered target animation frames. Note that both source and target local patches are extracted during the training stage, while only source local patches are extracted at the inference stage.

For the source video frames, we utilize an off-the-shelf facial landmark detection method, HRNet~\cite{wang2020deep}, to estimate 2D landmarks. These are depicted as red points in Figure~\ref{fig:infer3} (b). For alignment in each frame, the module calculates the center point of each eye and that of the mouth by determining the mean coordinates of the landmarks surrounding these three face elements. This process is unique to the source stage, where, regardless of the user's selection of local patches, only the eyes and mouth are chosen as the elements for alignment.
These three center points, represented as blue points in Figure~\ref{fig:infer3} (b), are then utilized to calculate the ultimate center point of the source face to align each frame into a canonical position and orientation. After alignment, an initial cropping operation isolates just the facial area from the video frame, eliminating unnecessary information. Next, a bounding box is defined for each face element in such a way that it encloses the full range of motion (ROM) around the current face element of focus, in this case the mouth, as shown in Figure~\ref{fig:infer3} (b). The local patches are then extracted by cropping and resizing the bounding box to $128\times128$ pixels. 

For the target 3D character, the process begins with extracting 2D landmarks for all target frames. For this, the user selects 3D vertices on the target mesh, whose orthographic projection will determine the 2D landmark positions.
Next, in the alignment step, a transformation matrix is calculated for each of the N face elements to align the target face element shape with that of the source. This calculation is based on the 2D landmarks of each face element in the target's anchor frame - a frame for which the rig parameters are set to zero - and the landmarks of the corresponding element in the first frame of the source video, which displays a neutral expression. The aligned target frames are then initialized by the bounding box parameters of the source. These are further adjusted in terms of width and height to sufficiently enclose the target's ROM. This is performed by increasing or decreasing the width and height to account for any size discrepancies between the target's ROM and the initialized bounding box. 

The per-frame landmark prediction results are derived solely to determine the ROM from the source data, instead of for precise frame-by-frame alignment. Consequently, minor jittering in the landmark predictions does not affect the overall aligning process, as the primary purpose is to estimate the maximum extents of facial movements. Finally, each of the aligned N target face elements are cropped and resized according to these element-specific bounding boxes. Consequently, a local patch of size 128x128 is obtained for each of the N face elements. Figure~\ref{fig:infer3} (b) demonstrates this process per face element basis, specifically for the mouth.
The decision on which local patches to use is informed by the most salient motions observed in the animation data. Common areas of focus include around the eyes and lips, although other areas such as the forehead, chin, and nose can also be considered if they exhibit significant motion.


\begin{figure}[!t]
    \centerline{\includegraphics[width=\linewidth]{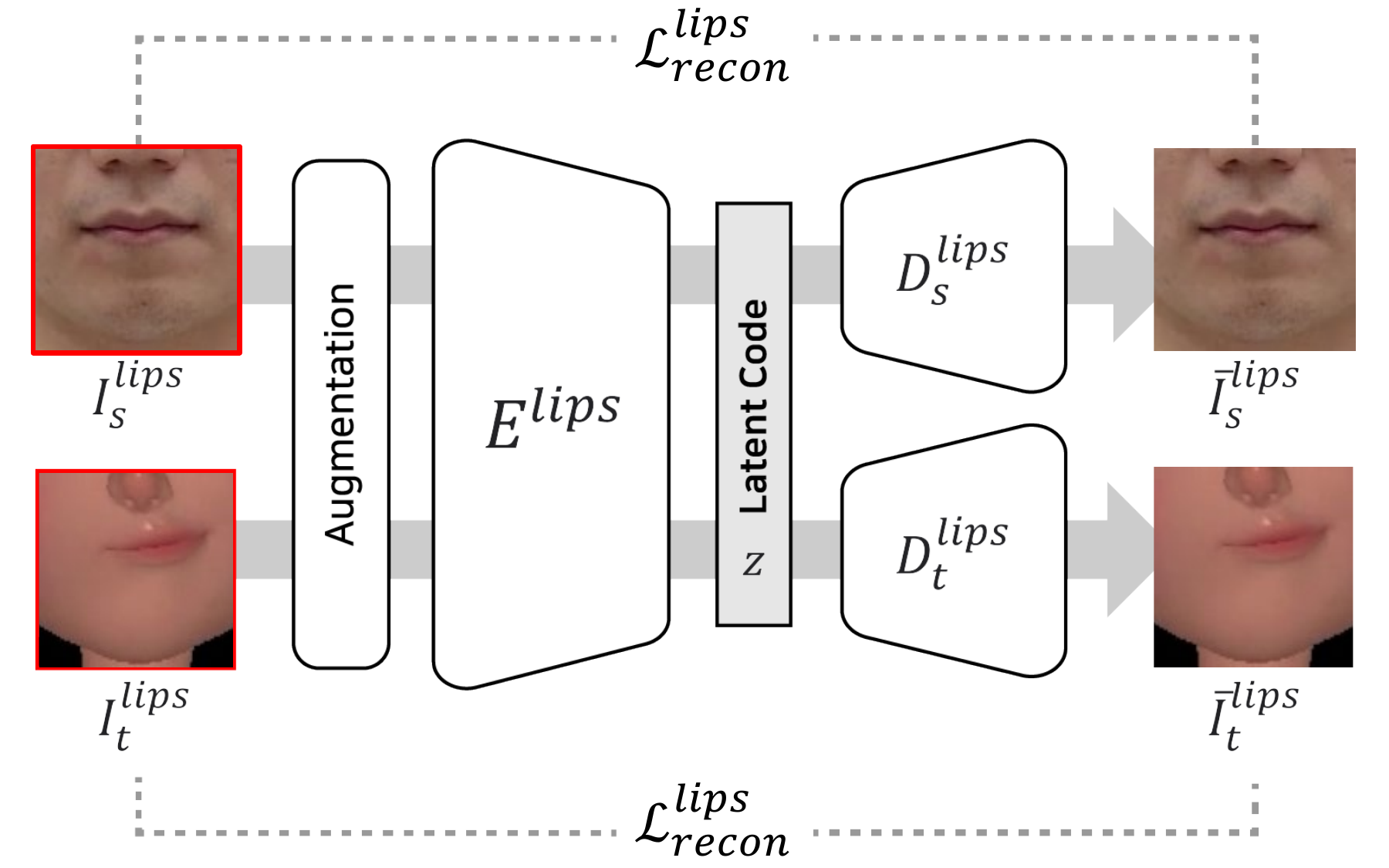}}
    \caption{ 
    The training process of the autoencoder for a local patch in RM. (The local patch is extracted from lips as an example.)
    An image reconstruction loss $\mathcal{L}_{\text {recon}}^{lips}$ is applied to train decoders $D_{s}^{lips}$ and $D_{t}^{lips}$ to reconstruct the source patch $\bar{I}_{s}^{lips}$ and the target patch $\bar{I}_{t}^{lips}$, respectively. 
    } \label{fig:REM}
\end{figure}

\subsection{Reenactment Module} \label{REM}

RM applies image-to-image translation to given source local patches to generate reenacted target local patches. 
The module can have any number of autoencoders corresponding to the number of local patches used. 
The autoencoders share the same architecture, which is identical to that proposed in Moser et al.~\cite{moser2021semi}. Figure~\ref{fig:REM} illustrates an overview of training one of the autoencoders.  
Each autoencoder consists of one encoder $E^{L}$ and two decoders $D_{s}^{L}$ and $D_{t}^{L}$ for the source and the target, respectively.
Note that $L$ is a local patch for the face element used in the retargeting process ($L \in \left\{ \text{left eye, right eye, lips, and etc.} \right\}$)

The local patches of the source $I_{s}^{L}$ and of the target $I_{t}^{L}$ are encoded into a common latent space $Z$ using a shared encoder $E^{L}$. The latent code $z \in Z$ is then fed to both decoders $D_{s}^{L}$ and $D_{t}^{L}$ to produce $\bar{I}_{s}^{L}$, $\bar{I}_{t}^{L}$, respectively. 
The autoencoders are trained in an unsupervised manner, requiring only input data for reconstructing both source and target, without the need for paired source and target image data. 
The networks are trained with the following two loss functions.

$L_1$ pixel-wise distance is calculated between the ground truth and the generated patch for both the source and target. The loss function is expressed as follows:
\begin{equation}\label{recon_L}
\begin{split}
    \mathcal{L}_{L_1}^{L}=\left\|I_s^{L}-\bar{I}_s^{L}\right\|_1+\left\|I_t^{L}-\bar{I}_t^{L}\right\|_1,
\end{split}
\end{equation}
where $I_s^{L}$ is the source local patch, $I_t^{L}$ is the target local patch, $\bar{I}_s^{L}$ is the generated local patch for source, and $\bar{I}_t^{L}$ is the generated local patch for target.

A structural similarity is also calculated between the ground truth and the generated patch for both source and the target. The loss function is expressed as follows:
\begin{equation}\label{recon_SSIM}
\begin{split}
    \mathcal{L}_{\text {SSIM}}^{L}=\text {SSIM}\left(I_s^{L}, \bar{I}_s^{L}\right)+\text {SSIM}\left(I_t^{L}, \bar{I}_t^{L}\right),
\end{split}
\end{equation}
where $SSIM(\cdot)$ is the structural similarity index measure as proposed in Wang et al.~\cite{wang2004image}.

The total loss function for the autoencoder is expressed as follows:
\begin{equation}\label{ae}
\begin{split}
    \mathcal{L}_{recon}^{L}=\mathcal{L}_{L_1}^{L}+\mathcal{L}_{\text {SSIM}}^{L}.
\end{split}
\end{equation}
The autoencoders in RM are jointly trained. After the training, the encoder and the target decoder for each local patch are used in the inference step.

\begin{figure}[ht]
    \centerline{\includegraphics[width=\linewidth]{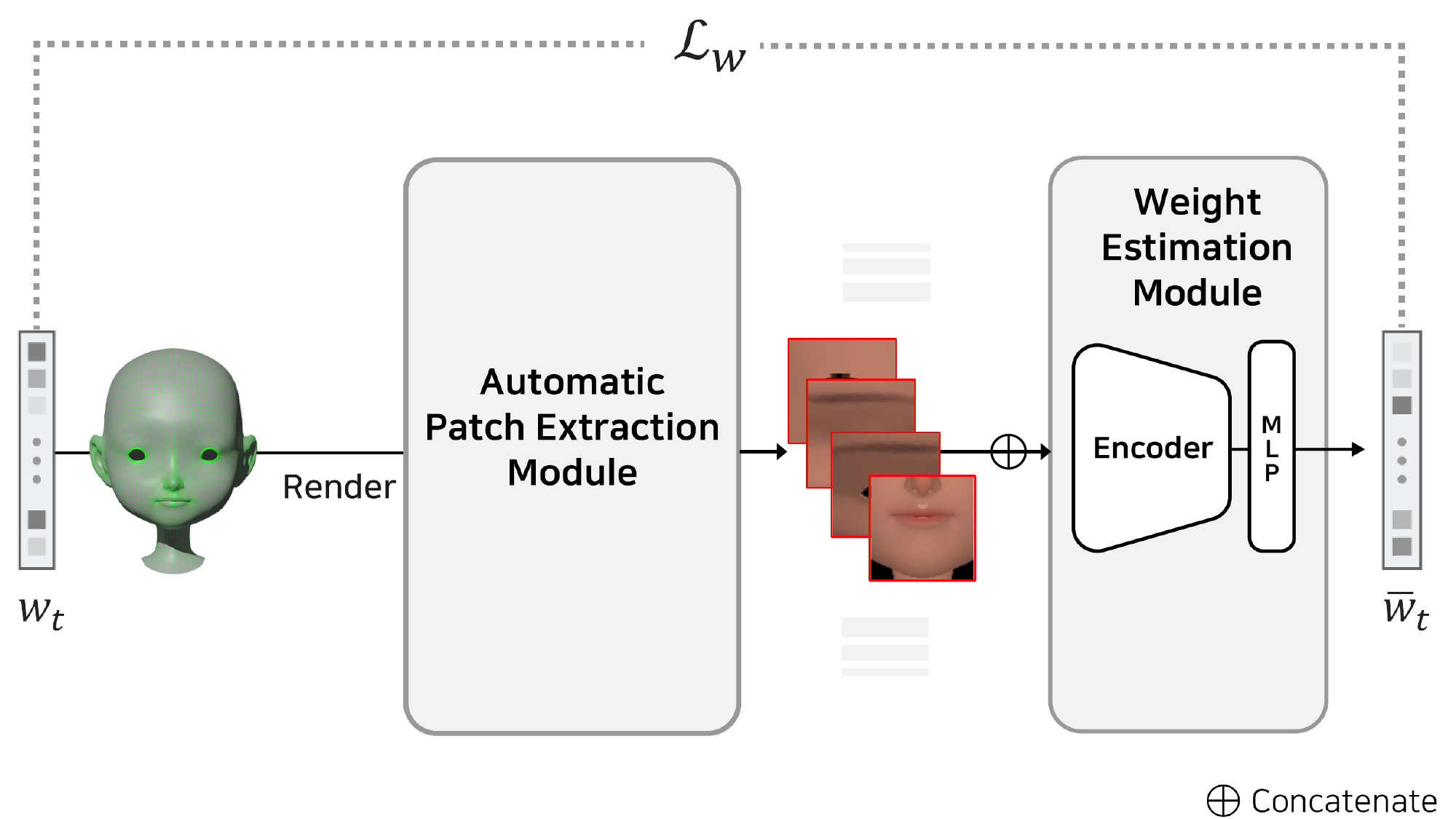}}
    \caption[Weight Estimation Module]{ 
    The training process of WEM. WEM estimates PCA weights $\bar{w}_t$ from the local patches of the target character. $\mathcal{L}_w$ compares the difference between estimated $\bar{w}_t$ and $w_t$ that was acquired to construct the animation sequence of the target character.
    } \label{fig:WEM}
\end{figure}


\begin{figure}[!t]
    \centerline{\includegraphics[width=\linewidth]{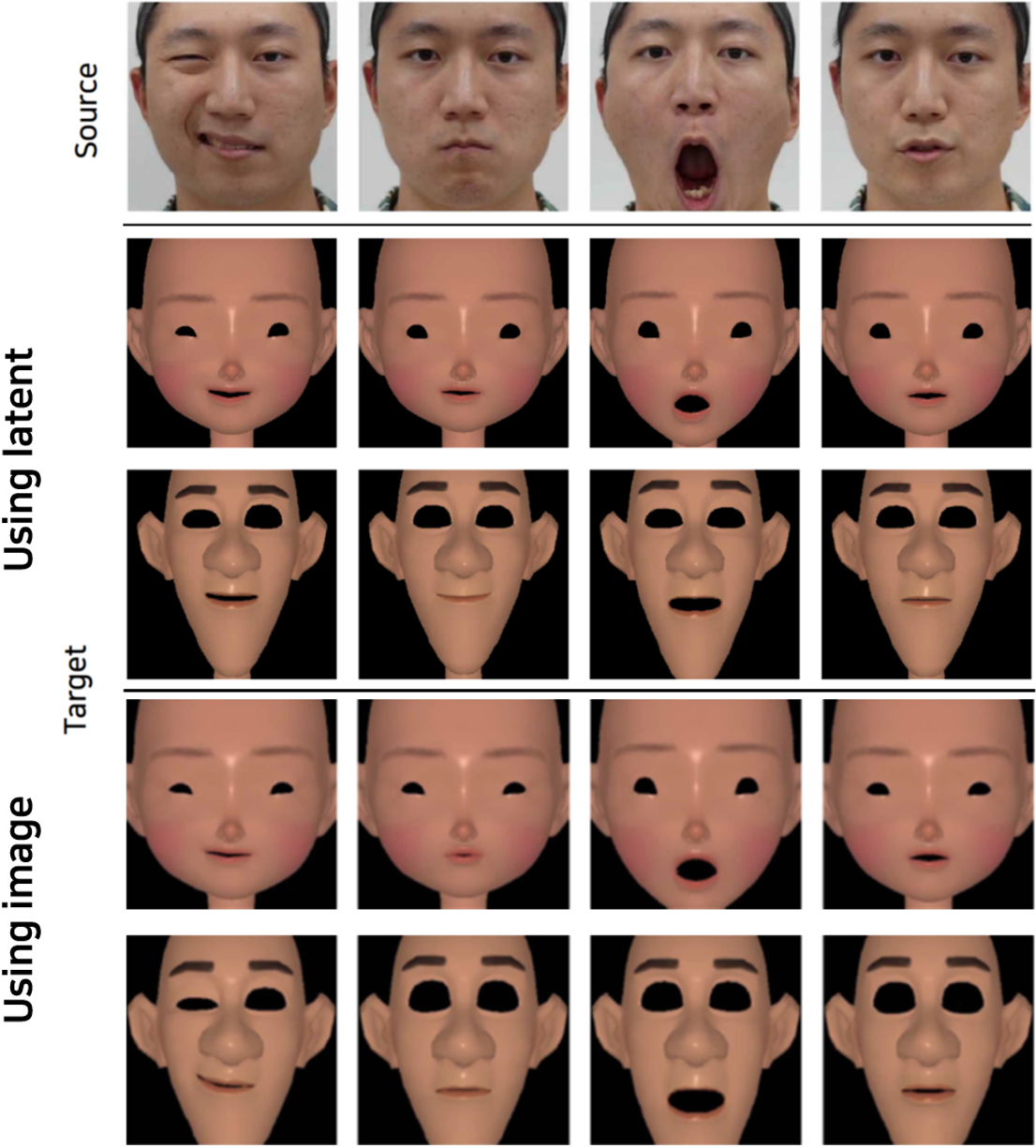}}
    \caption{
    Evaluation of WEM training using the latent space directly versus using the images predicted by RM. 
    }\label{fig:latent}
\end{figure}

\subsection{Weight Estimation Module} \label{WEM}

Given reenacted target local patches, WEM estimates the PCA parameters for 3D facial animation. WEM consists of one encoder which encodes the given local patches into a latent vector and one MLP that estimates the scalar weights for the PCA coefficients from the encoded latent vector.  
For the input representation of the network, we experimented with two setups: 1) using images predicted by RM, and 2) directly using the latent space learned by RM. As shown in Figure~\ref{fig:latent}, the result indicates that using the predicted images produces more accurately retargeted expressions than using the latent space directly. Inputting local patch images enables WEM to estimate parameters exclusively from the target data, whereas using the latent space requires estimating parameters from the encoded source and target information, adding complexity. Leveraging the learned latent expression space would also increase training time, as WEM would need to be trained after RM is fully trained, introducing a sequential dependency. As a result, unlike Moser et al., we chose to input local patches in the image space instead of directly using encoded latent vectors.

The local patches from RM are concatenated in a channel-wise manner before being fed to the encoder. Given this concatenated input, we employ a single global network architecture for WEM because it estimates the PCA parameters, derived from the entire animated mesh sequence. Because each PCA basis represents a significant global facial variation in descending order,instead of a semantically disentangled expression, it is not feasible to process the PCA parameters locally. The single encoder then projects the concatenated global input to its learned latent space to find a latent vector of dimension $2048$ in a similar way to Kim et al.~\cite{kim2021deep}. The vector is fed to the MLP to produce PCA weights.
To train the networks, we use a paired set of data: PCA weights, $w_t$, and the rendered local patches, $I_t^{L}$, of the target stylized 3D character obtained in Section~\ref{data preparation}. 
The training process is visualized in Figure~\ref{fig:WEM}.
We use $L_1$ loss to calculate the error between ground truth weight $w_t$ and estimated weight $\bar{w}_t$ through the following equation:
\begin{equation}\label{param}
    \mathcal{L}_w=\left\|w_t-\bar{w}_t\right\|_1.
\end{equation}

\begin{figure*}[ht]
    \centerline{\includegraphics[width=\linewidth]{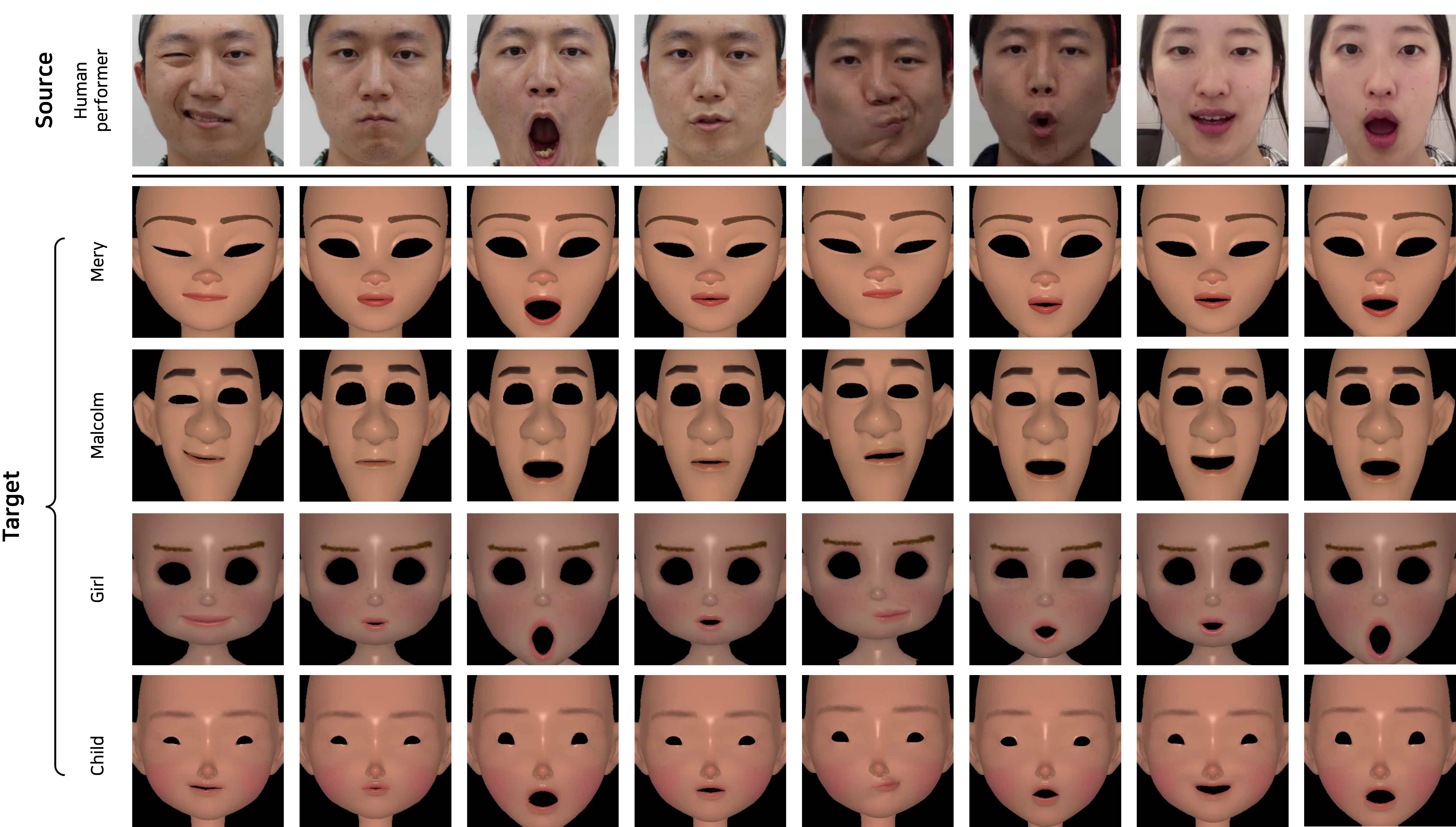}}
    \caption[ours]{Retargeting results from a human performance video to various 3D character models with different shapes and styles. As shown in Table~\ref{fig:data}, all of the above characters utilized three local patches. The columns one to six show results from a male performer and columns seven to eight show results from a female performer using unseen expressions. The source in the fifth and sixth columns was captured using the same male performer in a completely different environment from that used for capturing training data. \small{© Face rigs: meryproject.com, 2023 AnimSchool}
    } 
    \label{fig:ours}
\end{figure*}

\subsection{Inference}
After training the networks, the local patches are automatically extracted from each frame of the source video and are processed through RM and WEM sequentially to produce the PCA weights for each frame as shown in Figure~\ref{fig:infer3} (a). 
In the inference step, the encoder $E^{L}$ and the target decoders $D_t^{L}$ in RM are used for the reenactment process.
The PCA weights estimated from WEM are multiplied with the pre-calculated PCA basis to produce a facial animation for the target character.

The per-frame animation estimated by our method can be seamlessly exported to popular animation platforms such as Blender or Maya. This facilitates convenient integration into existing animation workflows. For editing and adaptation to different rig systems, the PCA parameters can be converted to other rig values through external post-processing. This process may involve solvers that align the estimated animations with feasible shapes generated by blendshapes and corrective shapes, as described in Moser et al.~\cite{moser2021semi}. This adaptability enhances the utility of the method in various animation production environments.

\begin{table}[h]
    \centering
    \caption[data]{Details of 3D character used for experiments. The columns show the number of vertices of each character, the number of frames used to train RM and WEM, and the number of patches used.
    } \label{fig:data}
    \includegraphics[width=1.0\linewidth]{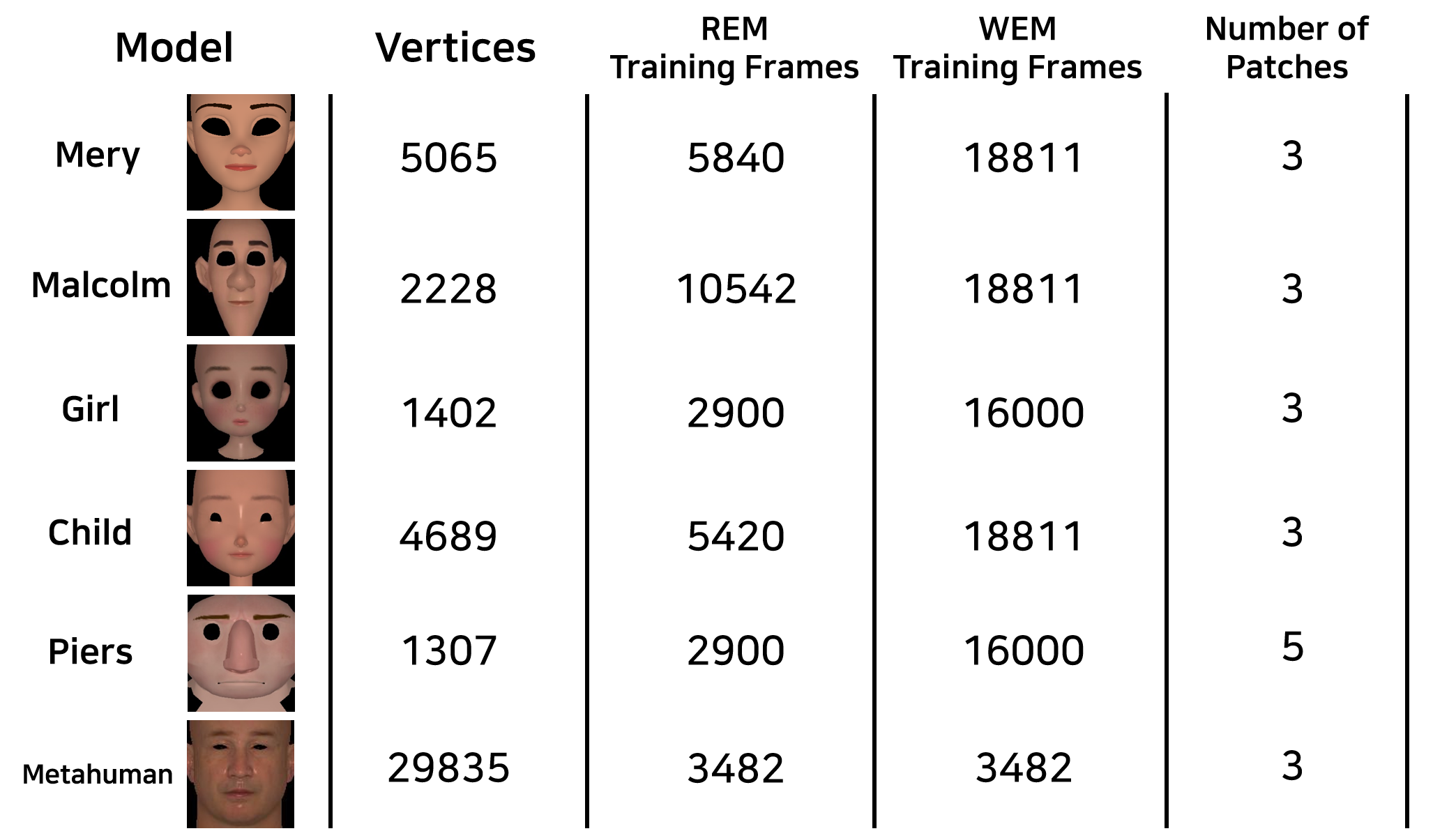}
\end{table}


\begin{figure*}[htp]
    \centerline{\includegraphics[width=.95\linewidth]{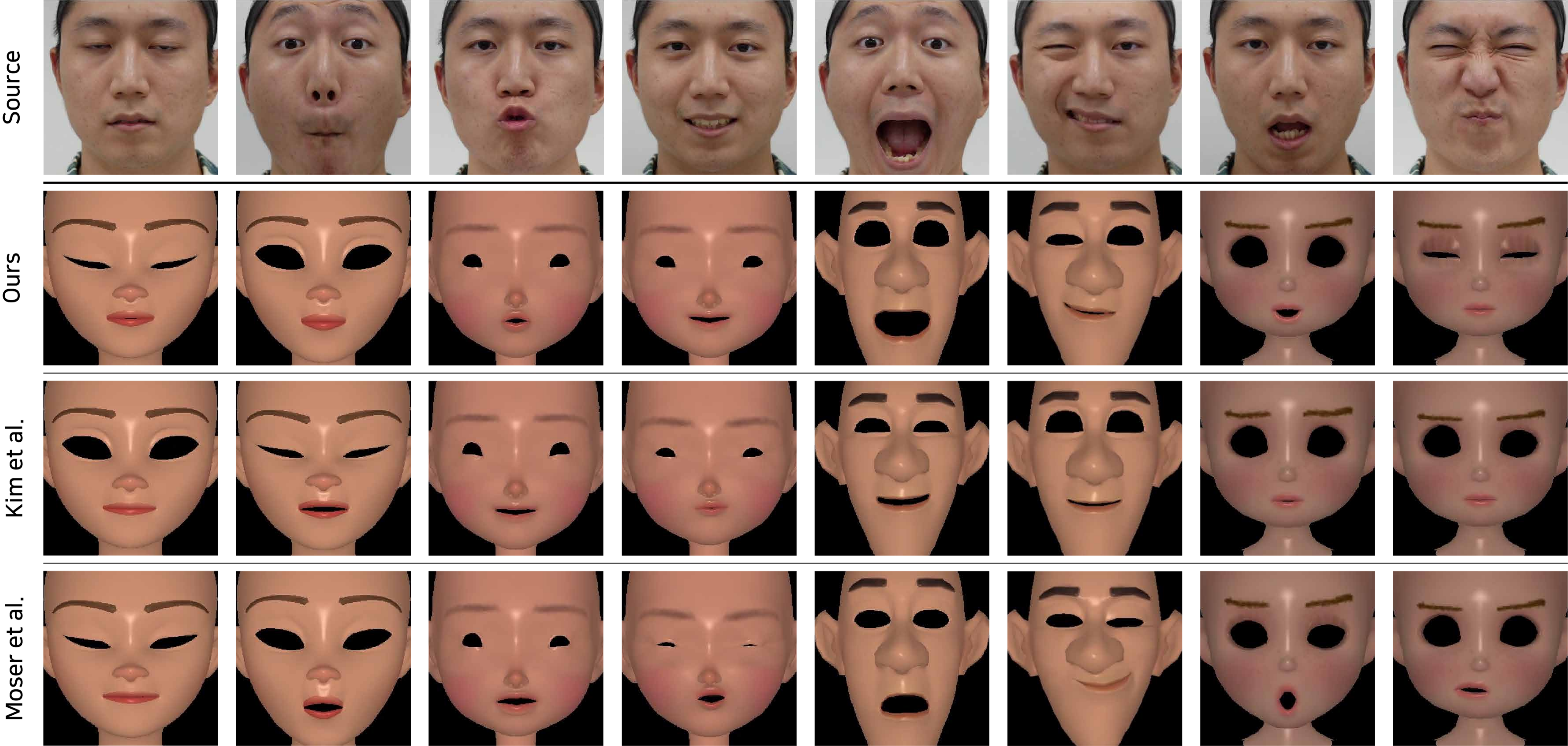}}
    \caption[Comparison with previous methods]{
    Comparison of retargeting results with previous methods using a video captured under the same lighting condition as that used for training. Results for each character (Mery, Child, Malcolm, Girl) are presented across two columns from left to right.
    \small{© Face rigs: meryproject.com, 2023 AnimSchool}
    } \label{fig:comparison}
\end{figure*}

\begin{table}[]
\caption{Training and inference time comparisons.}
\begin{tabularx}{\linewidth}{c *{2}{Y}}
\toprule
Methods & Train(m) & Inference(ms) \\
\midrule
Kim et al.     & 47 & 2.69 \\
Moser et al.  & 3904 & 6.31 \\
Ours        & 774 & 24.11 \\
\bottomrule
\end{tabularx}
\label{tab:trainTime}
\end{table}

\section{Experiments}

In this section, we describe implementation details and report the results of a comparison between the proposed method and previous methods. We also performed experiments to validate our design choices.

\subsection{Implementation Details}\label{implement}

We utilized five distinct stylized 3D characters to gather our animation data: Mery (© meryproject.com), Malcolm (© 2023 AnimSchool), Child, Girl, and Piers. The dataset comprised $18,811$ frames obtained using ARkit for Mery, Malcolm, and Child, and additional $16,000$ frames handcrafted by an artist for Girl and Piers. When training RM, the frame rate was adjusted to $24$ FPS, and the number of frames with neutral expressions were reduced to construct a dataset that has a balanced distribution of expressions. The facial movements of Mery, Malcolm, Child, and Girl were predominantly observed around the eyes, eyebrows, and mouth, leading to the use of three patches for these models. In contrast, Piers, exhibiting a distinct range of motion, particularly in the eye and eyebrow areas, necessitated the use of five separate patches to capture these nuances effectively. For the realistic character 'Metahuman' (© Metahuman), it was created by artists using the Metahuman Creator and utilized the same set of patches as Mery and other characters.

Table~\ref{fig:data} provides a detailed breakdown of these specifications. RM was trained for 50,000 iterations with a batch size of 16, using the Adam optimizer with a learning rate of $5 e^{-5}$. WEM was trained also for 50,000 iterations with a batch size of 64 using the same optimizer setting. Both modules were trained using local patches in a resolution of $128\times128\times3$, on a single consumer-level GPU, NVIDIA GeForce RTX 3090. We used PyTorch when implementing the architecture. Table~\ref{tab:trainTime} compares the total training and inference times for three different methods. The training time (in minutes) was estimated for the whole process, including reenactment and weight estimation for all methods. Our method required 774 minutes of training, which is longer than Kim et al. (47 minutes) but substantially shorter than Moser et al. (3904 minutes, as they trained each module for 50,000 iterations). However, in terms of inference time (in milliseconds), our approach is slightly slower than the others. Nonetheless, our method's 24.11ms inference time is close to real-time performance(13ms) for many practical applications.

\begin{figure*}[ht]
    \centerline{\includegraphics[width=.95\linewidth]{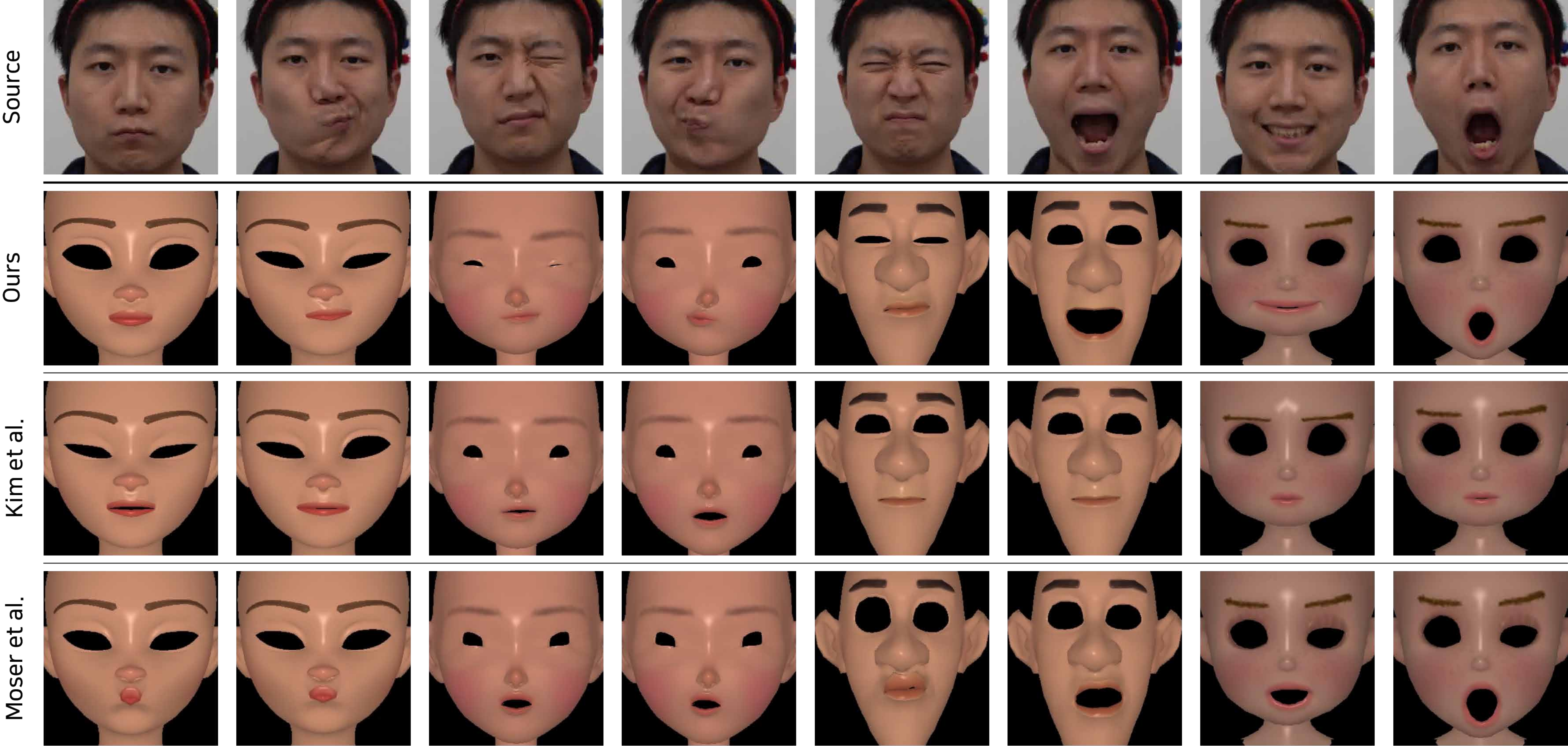}}
    \caption[Comparison with previous methods]{
    Comparison of retargeting results with previous methods using a video captured under a lighting condition different from that used for training. Results for each character (Mery, Child, Malcolm, Girl) are presented across two columns from left to right.
    \small{© Face rigs: meryproject.com, 2023 AnimSchool}
    } \label{fig:comparison2}
\end{figure*}

\begin{figure}[ht]
    \centerline{\includegraphics[width=\linewidth]{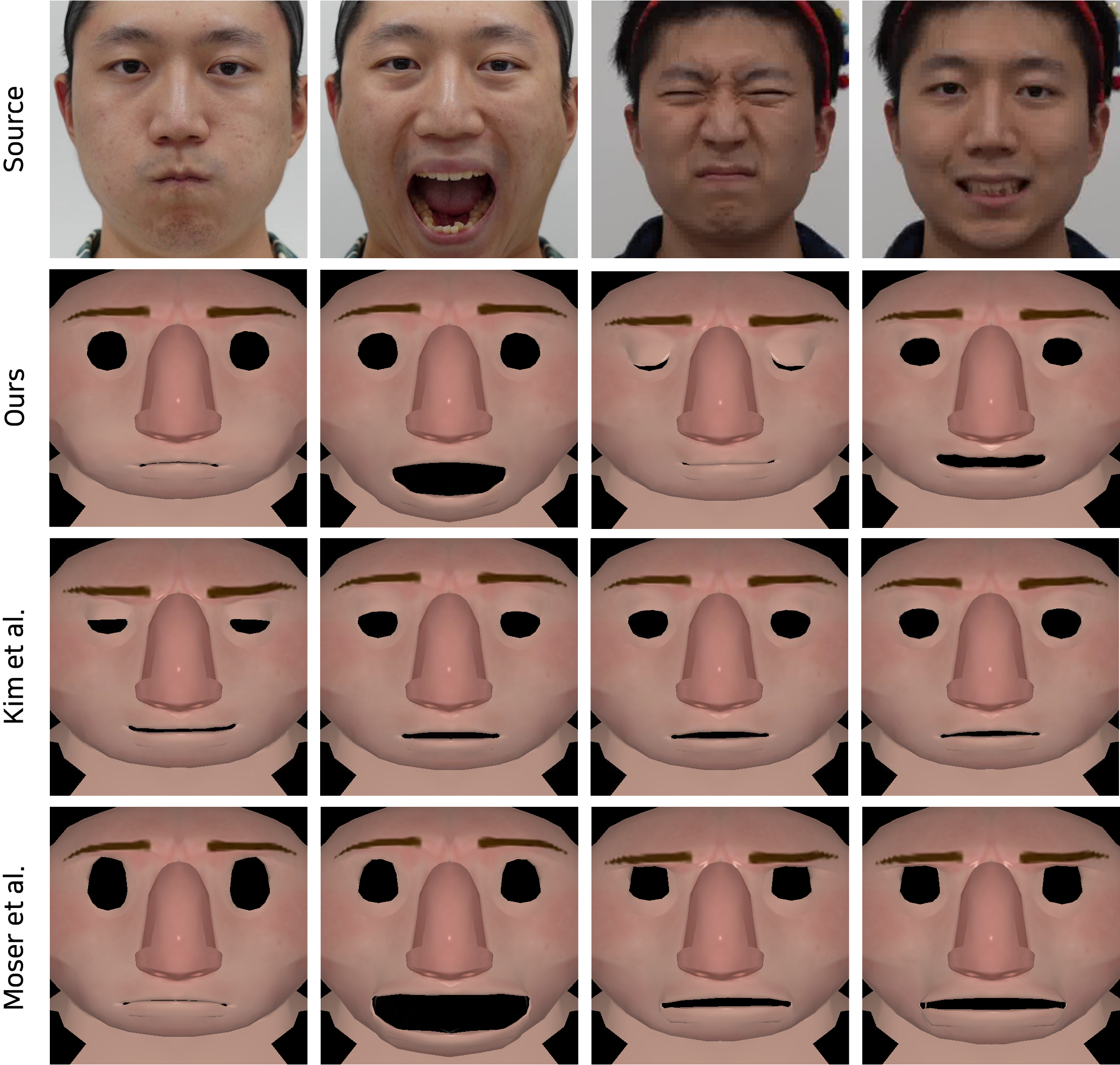}}
    \caption[Comparison with previous methods]{
    Comparison of retargeting results on Piers that employed five patches with previous methods using a video captured under the same lighting condition as that used for training(first two columns) and a lighting condition different from that used for training(last two columns).}
     \label{fig:comparison5p}
\end{figure}

\begin{figure}[ht]
    \centerline{\includegraphics[width=\linewidth]{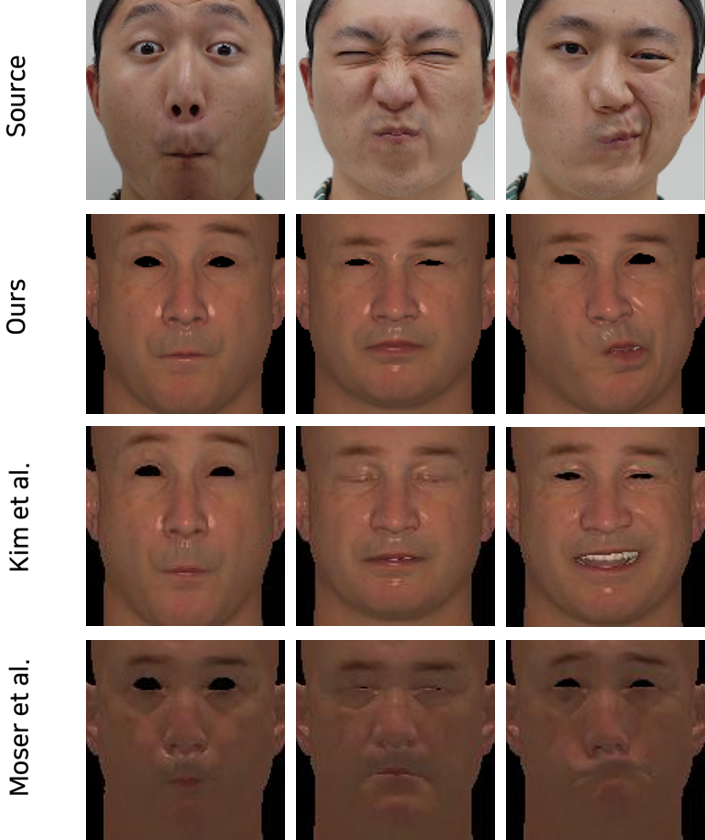}}
    \caption[Comparison with previous methods]{
    Comparison of retargeting results on realistic human characters.
     \small{© Metahuman}
     }
     \label{fig:realhumanComp}
\end{figure}

\subsection{Retargeting Results}

The retargeting results from the human performance captured in a video to the target stylized 3D characters are shown in Figure~\ref{fig:teaser} and Figure~\ref{fig:ours}. These results demonstrate the efficacy of our method in accurately conveying the expressions of the source performer across different stylized character models. Notably, even with significant proportional differences in facial features such as the eyes and mouth, the expressions are faithfully replicated on the target models. For instance, varied mouth shapes are reproduced well on Malcolm and Girl, whose face is much slimmer and wider, respectively, compared to that of the performers as shown in the second and third rows. In the sixth column, the accurately replicated puckered lips on Mery, Girl, and Child exemplify the precision of our method. For a more comprehensive view of our results, please refer to the supplementary video.

\begin{figure}[ht]
    \centering
    \centerline{\includegraphics[width=\linewidth]{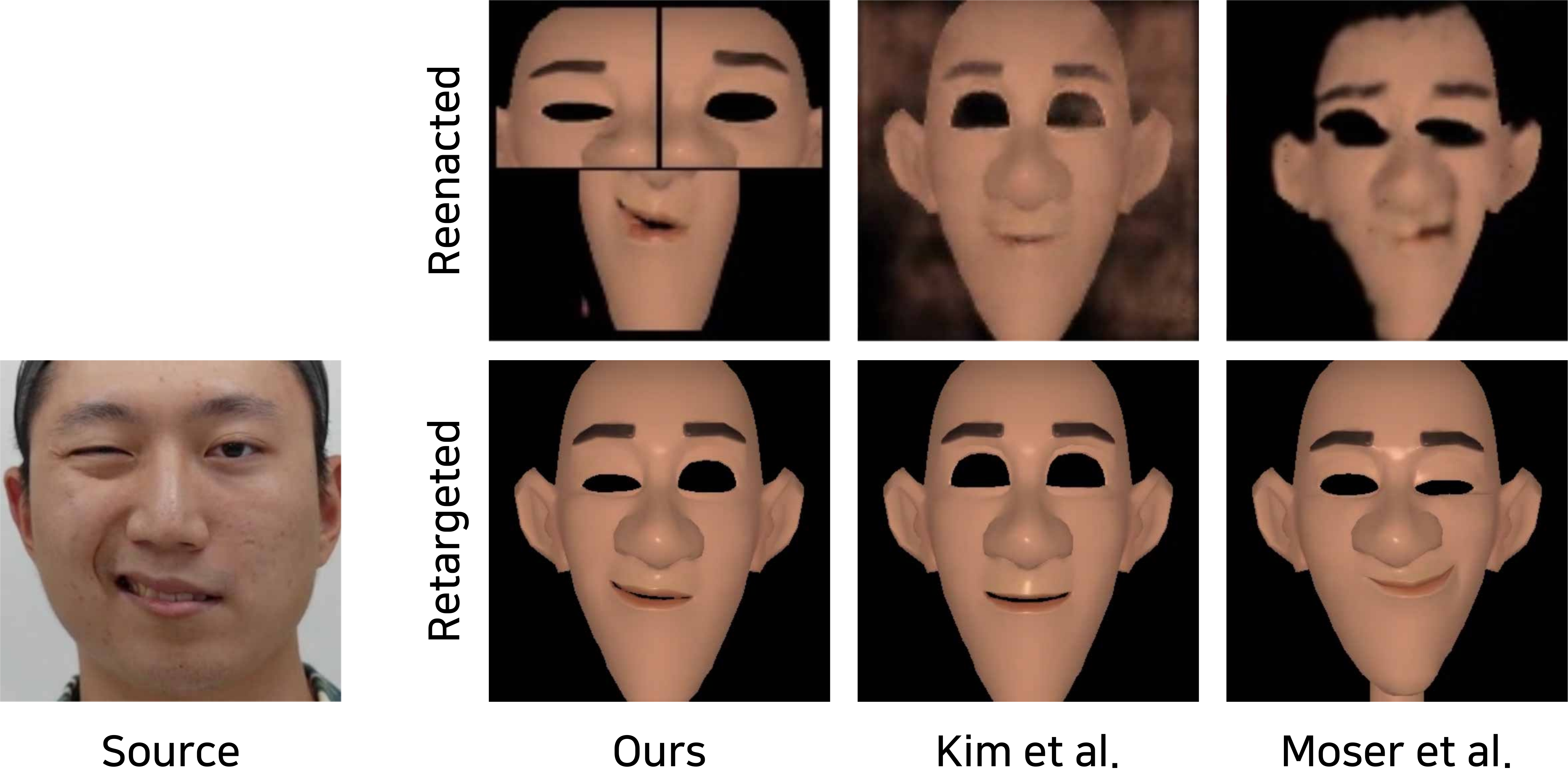}}
    \caption{
     Comparison of reenactment results. Unlike our method, inaccurate reenactment led to wrong retargeting results by previous methods. \small{© Face rigs: 2023 AnimSchool}
    } \label{fig:malcolm_reenact}
\end{figure}

\subsection{Comparisons}

In this section, we compared our method with two previous methods, Kim et al.~\cite{kim2021deep} and Moser et al.~\cite{moser2021semi} as demonstrated in Figure~\ref{fig:comparison},~\ref{fig:comparison2},~\ref{fig:comparison5p},~\ref{fig:realhumanComp} and in the supplementary video. 
We utilized the original implementation provided by Kim et al.~\cite{kim2021deep}. Because implementation of the work by Moser et al.~\cite{moser2021semi} was unavailable, we faithfully re-implemented the method based on the description provided in the original paper. All hyperparameters for each method were set as described in the respective papers. 

Initially, we conducted a comparison under the same lighting setting as that used for our training, as depicted in Figure~\ref{fig:comparison}. 
Figure~\ref{fig:comparison} displays examples of the result that show successful reproduction of facial movements on characters with much larger eyes than those of the performers, like Mery and Girl. 
The results produced by our method are semantically closer to the source expressions, as exemplified by a kissing expression (third column) and a wink expression (sixth column), compared to the results produced by previous methods. 

To evaluate the robustness of our method in practical scenarios involving variations in lighting conditions during facial capture of the source performer, we conducted an additional comparison under different lighting setting from those employed during the training. Despite being trained under a single light setting, our model exhibits remarkable resilience to lighting changes. This adaptability is evident in Figure~\ref{fig:comparison2}, where our model consistently produces high-quality results under darker lighting conditions, surpassing other methods in performance. This showcases the versatility of our approach.

Figure~\ref{fig:comparison5p} displays our retargeting results on the 'Piers' model, which utilized five patches, alongside those from previous methods. This demonstrates the capability of our method in retargeting complex expressions like cheek puffing, mouth opening, squinting, and smiling to models with significantly different facial shapes and sizes. It is apparent that the previous methods, which are designed to work on the entire face, struggle with intricate retargeting challenges that come from large differences in facial shapes and sizes. Figure~\ref{fig:realhumanComp} shows comparison results for realistic human characters. It demonstrates that our method can perform better than other methods in retargeting complex and diverse facial expressions for realistic human characters, such as pursing the lips, frowning, or twisting the mouth to the side.

Results of the reenactment of Malcolm produced by our method and by previous methods using the first expression used in Figure~\ref{fig:ours} are shown in Figure~\ref{fig:malcolm_reenact}.
The results in Figure~\ref{fig:malcolm_reenact} clearly indicate the superiority of our method over the previous methods in the retargeting performance. 
Because the reenactment results from the previous methods already contained severe artifacts due to the incapability of handling large structural differences between the source performer and the target character, the produced facial expressions did not match with the source expressions. By focusing on movements of the face element within local patches, our method was able to reenact the expressions from the source despite the differences in facial structure.

\begin{table}[]
\caption{Quantitative evaluation of character-to-character retargeting results.}
\begin{tabularx}{\linewidth}{c *{2}{Y}}
\toprule
\multirow{2}{*}{Methods} & \multicolumn{2}{c}{Vertex Error ($\times10^{-3}$ mm)} \\
\cmidrule(lr){2-3} & Mery$\rightarrow$Malcolm & Malcolm$\rightarrow$Mery \\
\midrule
\midrule
Kim et al.     & 1.4644 & 1.1305\\
Moser et al.  & 2.6543 & 6.3122 \\
Ours        & \textbf{1.4094} & \textbf{1.0587} \\
\bottomrule
\end{tabularx}
\label{tab:quan}
\end{table}

\begin{figure}[ht]
    \centerline{\includegraphics[width=\linewidth]{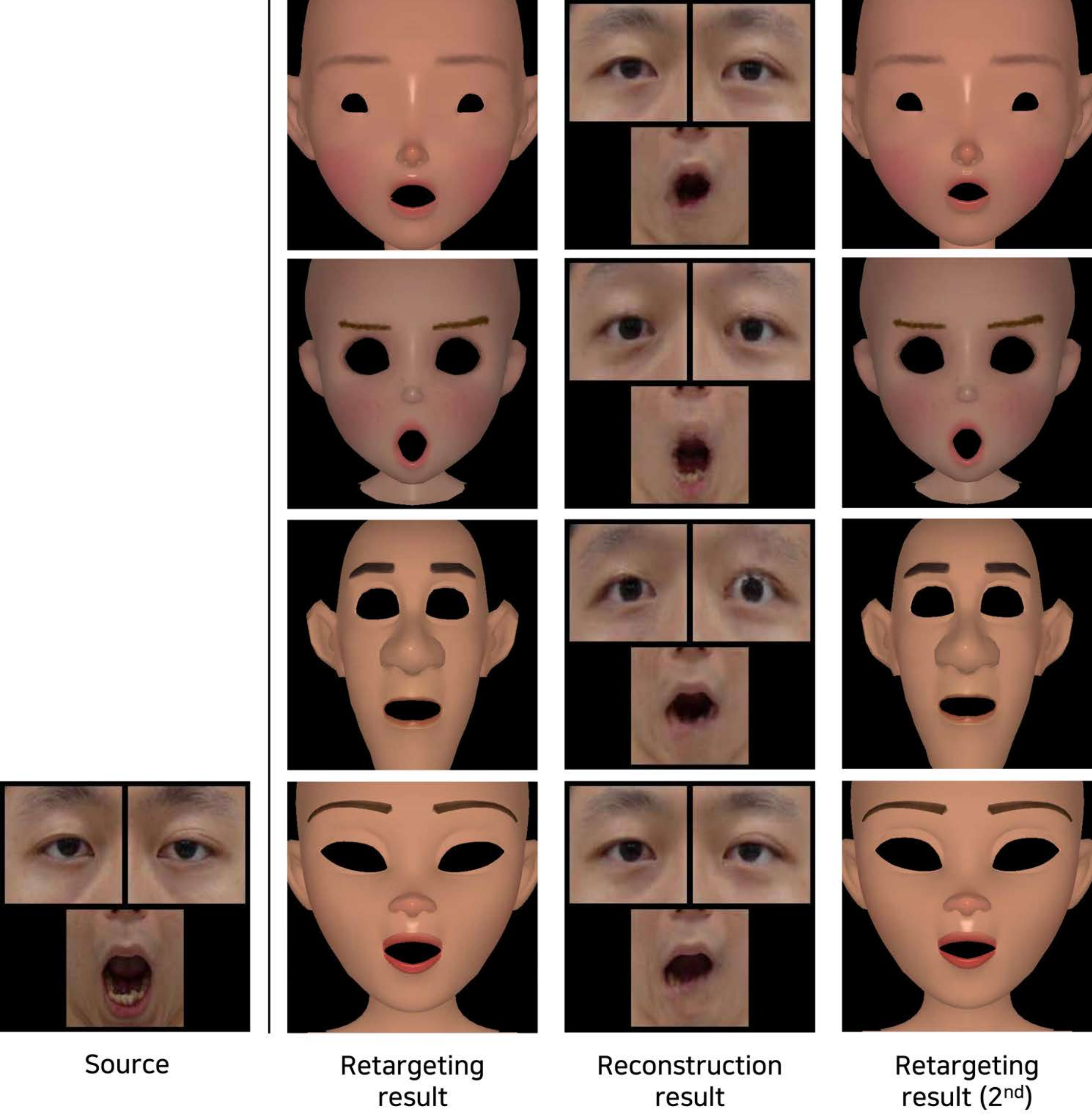}}
    \caption[Reconstruction]{Cyclic retargeting results. A source expression is retargeted to target character models. The retargeted image is then reenacted back to the source performer, which is again retargeted to the target character. \small{© Face rigs: meryproject.com, 2023 AnimSchool}
    } \label{fig:reconstruction}
\end{figure}

\begin{figure}[ht]
    \centerline{\includegraphics[width=\linewidth]{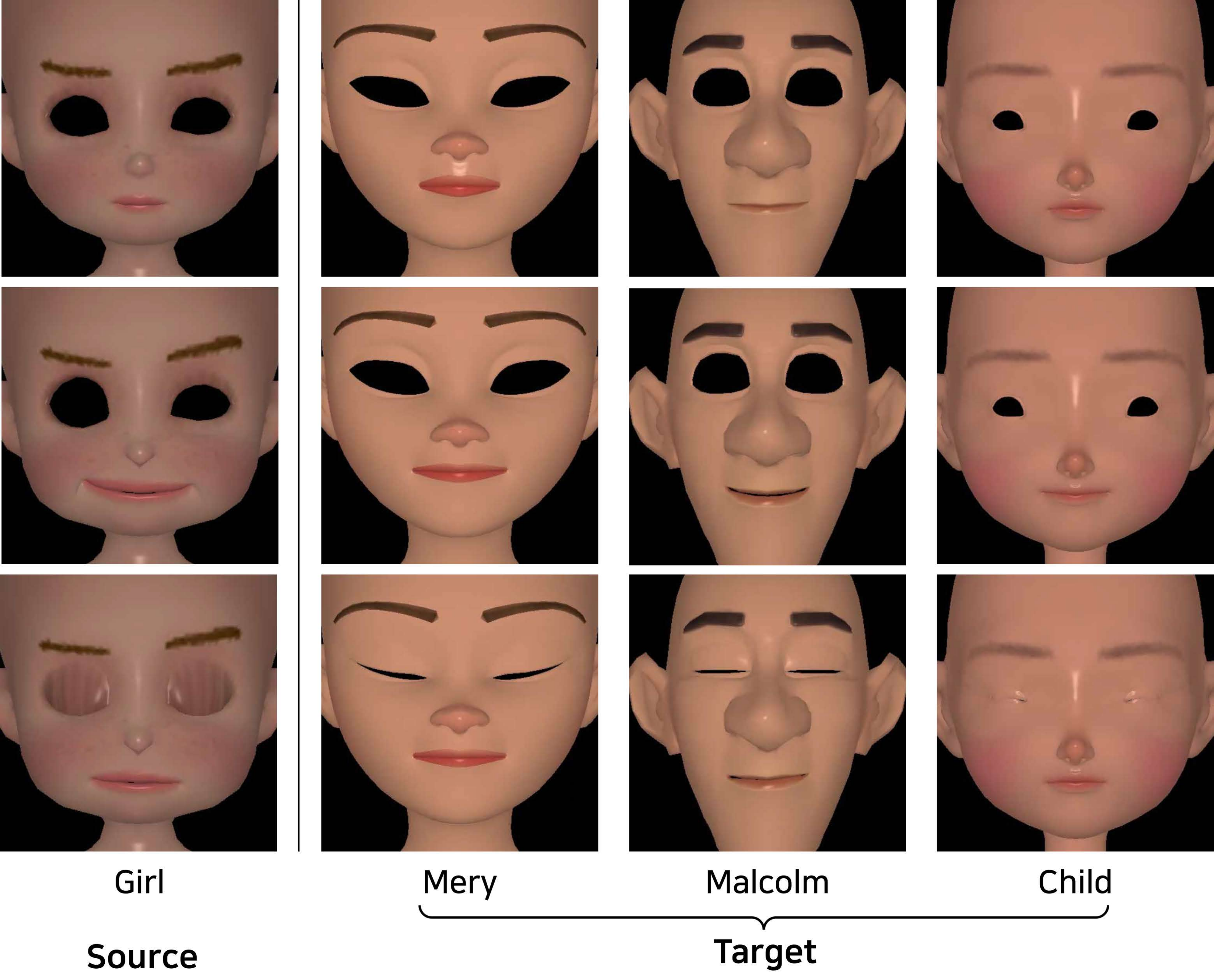}}
    \caption{Character-to-character retargeting results. \small{© Face rigs: meryproject.com, 2023 AnimSchool}
    } \label{fig:ch2ch}
\end{figure}

To quantitatively assess the retargeting quality of our method, we conducted a character-to-character retargeting experiment, as there are no ground truth animation parameters for the source human performer. Deviating from our initial configuration, we utilized rendered images of the character as the source and retargeted the character's animation to the target character. The experiment was carried out with Mery and Malcolm, which share the same WEM animation, as depicted in Table~\ref{fig:data}.
We employed a rendering of Mery, animated with Mery's WEM training data, as input and retargeted Mery's facial expressions to Malcolm. The resulting retargeted animations were then compared to Malcolm's WEM training data as the ground truth, because both characters share the same animation. This reverse process was also performed, retargeting Malcolm's expressions to Mery.
We computed the Mean Absolute Error (MAE) between the vertices of the retargeted target character and the corresponding ground truth. As shown in Table~\ref{tab:quan}, our method produced lower errors compared to other methods~\cite{kim2021deep}~\cite{moser2021semi}, indicating superior performance in accurately retargeting facial expressions.

\begin{figure}[ht]
    \centering
    \centerline{\includegraphics[width=0.9\linewidth]{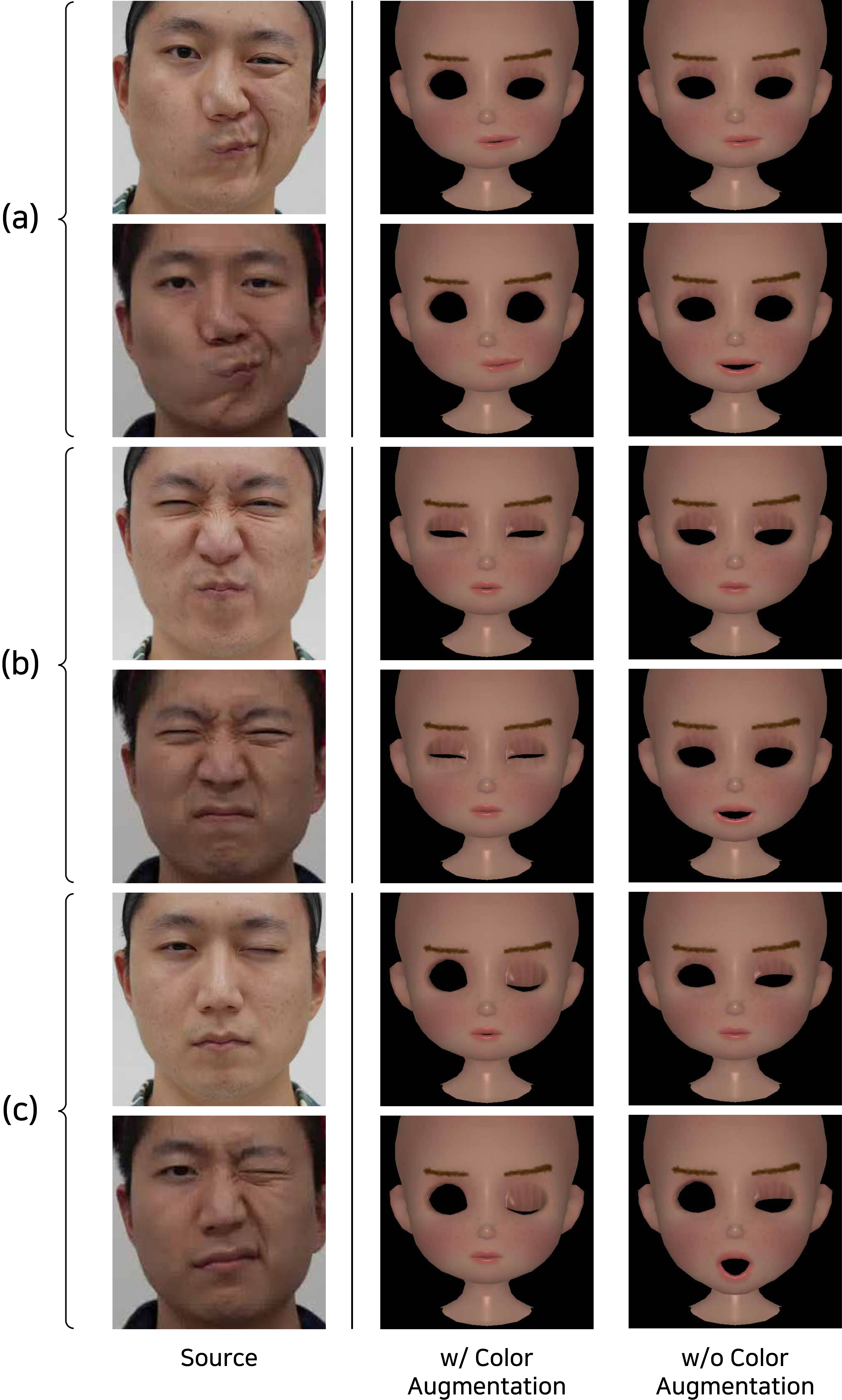}}
    \caption{
    Comparison of retargeting results produced with and without color augmentation. The source expressions are consistently reenacted for two similar frames under varying lighting conditions with color augmentation.
    } \label{fig:aug}
\end{figure}

\begin{figure}[h]
    \centering
    \centerline{\includegraphics[width=\linewidth]{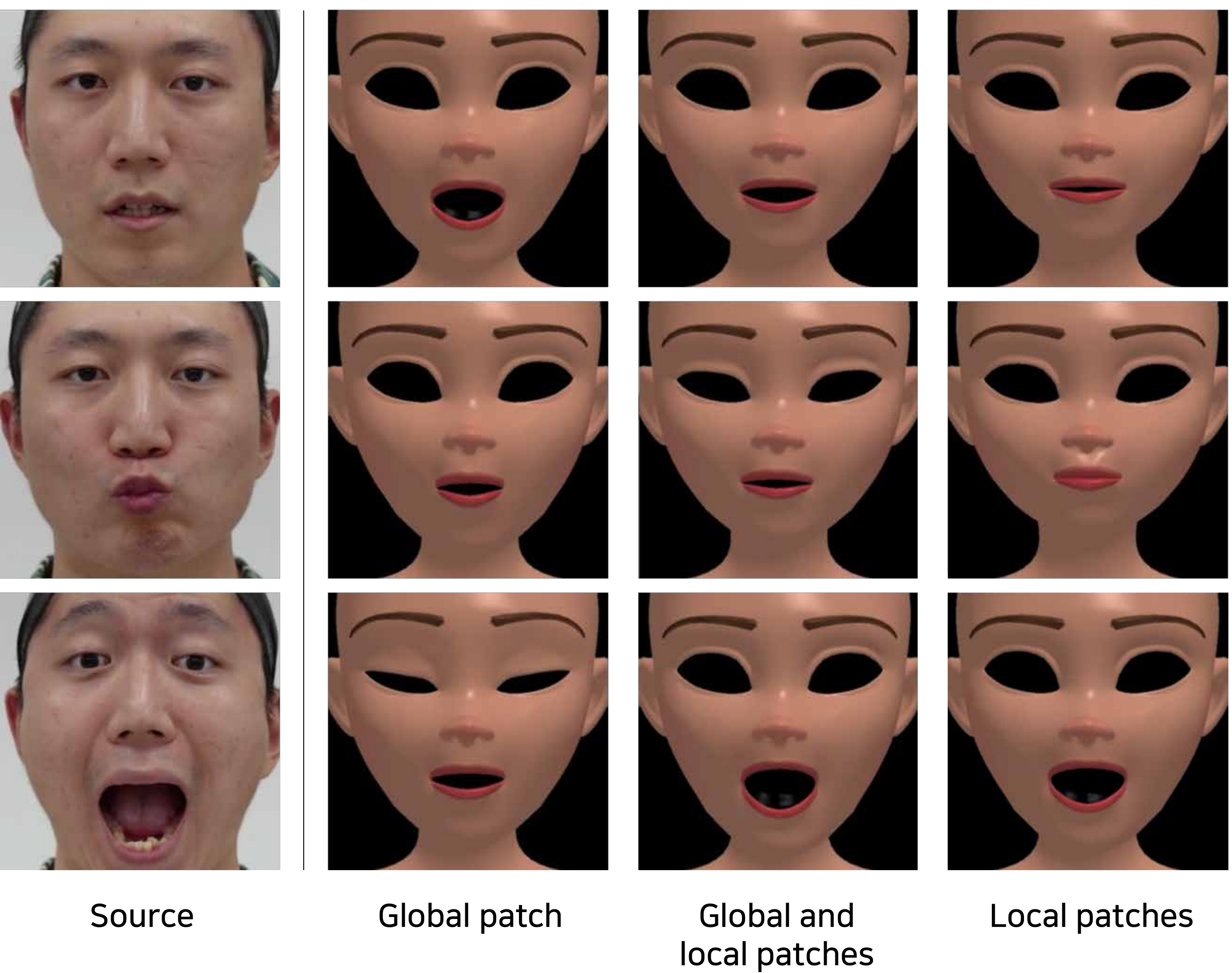}}
    \caption{Retargeting results produced using local and global patches. The best results are produced when only local patches are used. 
    \small{© Face rigs: meryproject.com}
    } \label{fig:input_format}
\end{figure}

\subsection{Cyclic Retargeting Results}

To check if there is information loss through our pipeline, we performed a cyclic retargeting. 
We first retargeted the facial animation captured from a human performer to four character models that employed three patches. We then rendered the retargeted results and used them as input to RM to reconstruct the original source human patches using the source decoder, $D_{s}^{L}$. Next, we used the reconstructed patches again as input to retarget the facial animation back to the character models. The cyclic retargeting results are shown in Figure~\ref{fig:reconstruction} and in the supplementary video. Ideally, the reconstructed source patch should be the same as the input patch, and the results from the second cycle should be the same as the initial retargeting results.
The details in the reconstructed source patches, such as hair and wrinkles, became slightly blurry but the movement of the face element captured in the local patch was well preserved, producing the results after the cyclic retargeting still appearing very close to the original facial expression.
This result verifies that our method can semantically match facial expressions between the source and the target and decode them to the desired local patches without losing much information through the cycle.

\subsection{Character-to-character Retargeting Results}

We performed additional experiments of character-to-character retargeting to further evaluate the generalizability of our method.
Our image-based approach enables us to use not only human performer videos as input but also the rendering of 3D character animations. We used the rendering of a 3D character animation as input and retargeted the facial expressions of one 3D character to another. Because we execute motion transfer on rendered 2D image space, the mesh structures between the source and target 3D models do not have to be identical, enabling mesh-agnostic retargeting. We set the rendering of the animation of Girl as the source video and retargeted it to other characters. The results are shown in Figure~\ref{fig:ch2ch} and in the supplementary video. The eye movements of Girl were properly reproduced on Mery and Child.
These results verify that our method is general enough to utilize different types of driving source: either video or 3D animation.

\subsection{Effect of Color Augmentation}\label{eval_color}

Color augmentation aims to ensure robust performance under changes in color tone or lighting conditions. To evaluate its effect, we trained RM with and without color augmentation and used two types of test videos captured from a single performer: one captured under the same lighting condition as that used for training data, and the other captured under a different lighting condition. Three similar facial expressions observed in both videos are compared in Figure~\ref{fig:aug}.
The facial expressions of the source videos observed in (b) and (c) are almost the same, while those observed in (a) are slightly different in the degree of closure of the right eye. 
The results produced by RM trained with color augmentation show semantically consistent facial expressions across similar frames under varying lighting conditions. In contrast, the states of mouth closure changed for all three expressions when the module was trained without color augmentation.

\subsection{Effect of Using Local Patch}

We experimented to evaluate the effect of using local patches. The experiment was conducted in three cases; using a global patch, which contains the entire face, using a global patch and local patches, and using only local patches. For both the global and local patch setups, we resized all individual patches to a resolution of 128x128, and then spatially concatenated them into a single composite image. 
As shown in Figure~\ref{fig:input_format}, the results of using a global patch were identical to those of previous image-based approaches~\cite{kim2021deep,moser2021semi}. 
When the global and local patches were used, the retargeting results were still unsatisfactory due to conflicting information between reenacted patches. We observed that the best option was to use only local patches as shown in Figure~\ref{fig:input_format}.

\vspace{-2mm}

\section{Discussion}

\subsection{Limitations}
Despite the improvement in retargeting facial animation from a human performer to a stylized 3D character, we acknowledge several limitations of our method. Our approach is effective for human-like characters with exaggerated facial features, but certain limitations arise with non-human characters with unique anatomical features. For example, features like an elephant’s trunk or a bug's antennae can be very challenging. These features fall outside our method’s scope, though we still account for most of the shape and size differences between the source and target.
Our method learns a shared representation between the source and target identities and decodes this information using an identity-specific decoder in an unpaired data setting. For our training to be successful, it is important to use animation data that contains a sufficient ROM. If the data do not comprehensively represent the ROM, the network may fail to establish an appropriate shared representation. These limitations offer avenues for future improvements and underscore the importance of rigorous data collection and preparation to enhance the performance of our facial animation retargeting method. Similarly, our method is inherently dependent on the target character's rig setup. Fine-grained details, such as wrinkles, are transferred only if the target's rig setup is capable of depicting such features. To achieve the transfer of these intricate details, a more sophisticated rig setup is necessary.

\begin{figure}[h]
    \centering
    \centerline{\includegraphics[width=\linewidth]{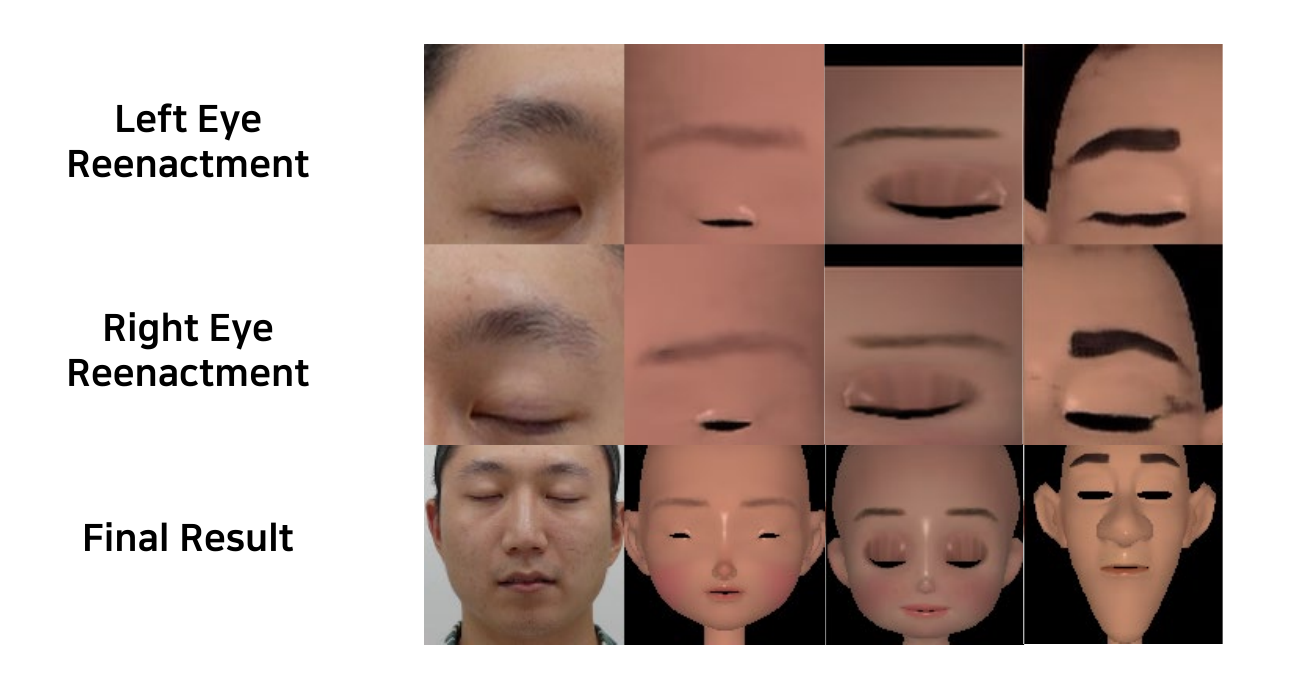}}
    \caption{The reenactment process and final retargeting results for the eye closing motion on the character child, girl, and Malcolm.
    } \label{fig:eye_closing}
\end{figure}

Through our analysis, as shown in Figure~\ref{fig:eye_closing}, we found that the full eye closing motion was not completely transferred due to an inherent limitation of the image-based approach. Specifically, when the human performer's eyes close completely as shown in the first and second rows, RM mistakenly aligned and recognized the eyelashes as part of the hollow black background, istead of as a fully closed eye. Consequently, as shown in the third row, RM interpreted the black eyelash region as a slight opening, which prevented the eyes from completely closing in the final WEM prediction result. This issue could potentially be addressed through advanced target modeling, such as adding eyelashes to the character mesh. Additionally, our method does not correctly transfer the motion for the yaw and pitch directions, while the transfer of roll motion was successful due to our data augmentation with randomly applied in-plane affine transformations.

\vspace{-2mm}

\subsection{Future work}

The proposed APEM requires the user to define the vertices on the target 3D character model to obtain 2D landmarks for local patch extraction.
An interesting future research direction would be developing an automatic landmark detection technique for any type of stylized 3D characters that can help select the face region with dominant movements by analyzing the given facial animation. 
This extension would allow a possible research direction for the automatic identification of the optimal local patch based on the target character's ROM. 
The alignment between the corresponding local patches is crucial in image-based reenactment methods, and any disparity can lead to significant quality degradation in the output. Therefore, another possible future work is to explore the application of methods that align images into canonical coordinates~\cite{peebles2022gan, yang2022dense, ofri2023neural}.

Our method currently does not account for retargeting variations in head poses, which presents another potential area for future research. Incorporating head pose as an additional condition could enrich the realism and applicability of the method. Although the proposed modules generalize sufficiently well across various lighting conditions, our main focus was the successful transfer of facial expressions from a human performer to a single stylized character. Consequently, they are designed to deal with a single performer per trained model for each target character. One way to expand the capability of our model is to use a pretrained 2D prior, such as StyleGAN~\cite{karras2019stylebased}, for neural expression feature extraction from various identities. Utilizing these features could significantly improve the model’s ability to generalize across multiple performers.

\section{Conclusions}
In this paper, we introduce a novel local patch-based retargeting method designed to transfer facial animations from a human performer to stylized 3D characters, accommodating variations in face element shapes and sizes. Central to our approach is the focus on individual face element movements during the expression transfer process. This is accomplished using APEM, which efficiently extracts and aligns local patches from both source and target in the image space during training, thereby enabling expression retargeting at inference despite structural disparities.
Our method incorporates three primary modules: the aforementioned APEM for patch extraction; RM, which replicates the target's local patches based on the source's patches; and WEM, responsible for calculating PCA weights from the reenacted target patches. Together, these modules enable the production of facial animations for the target 3D character, transferring expressions patch by patch, and frame by frame from the source video. 
It is important to note that our main focus lies in the novel ‘local’ patch extraction, in which the accurate alignment between the source and target faces via patch-based architecture is ensured even with large shape differences of face elements. Given only the vertices selected by the user, the alignment between the source and target patches is automatically performed through the newly proposed APEM process.
Our results demonstrate that this method effectively transfers facial animations from live performers to various stylized 3D characters. Notably, it capitalizes on the full range of the target character's facial motions, ensuring a dynamic and authentic animation experience.


\printbibliography           

\section*{Data Availability Statement}
The data that support the findings of this study are openly available in the following repositories:

\begin{itemize}\setlength{\itemsep}{0pt}
    \item \textbf{Mery} - Available from the Mery Project at \url{https://www.meryproject.com}.
    \item \textbf{Malcolm} - Available from AnimSchool at \url{https://www.animschool.com}.
    \item \textbf{Child} - This dataset was created in-house by our research team.
    \item \textbf{Girl} - This dataset was downloaded for free from cgtrader at 
    \url{https://www.cgtrader.com/free-3d-models/character/child/cartoon-girl-rigged-578fe010-4c89-4724-96f3-de378cf64817}
    \item \textbf{Piers} - This dataset was downloaded for free from cgtrader at 
    \url{https://www.cgtrader.com/free-3d-models/character/man/maya-character-rig-piers-3d-rig}
    \item \textbf{Metahuman} - This dataset was created in-house by or research team using Metahuman creator.
    \url{https://www.unrealengine.com/en-US/metahuman}
\end{itemize}

\section*{Conflict of Interest Statement}
The authors declare that there are no conflicts of interest regarding the publication of this paper.

\end{document}